\documentclass[journal]{IEEEtran}
\usepackage{amsmath,amsfonts,amssymb}
\usepackage{algorithmic}
\usepackage{algorithm}
\usepackage{array}
\usepackage[caption=false,font=normalsize,labelfont=sf,textfont=sf]{subfig}
\usepackage{textcomp}
\usepackage{stfloats}
\usepackage{url}
\usepackage{verbatim}
\usepackage{graphicx}
\usepackage{cite}
\usepackage{booktabs}
\usepackage{multirow}
\usepackage{xcolor}
\usepackage{colortbl}
\usepackage{tikz}
\usepackage{pifont}
\usetikzlibrary{positioning, arrows.meta, shapes, calc}
\usepackage{tikz}
\usepackage[implicit=false]{hyperref}
\hypersetup{hidelinks,
	colorlinks=true,
	allcolors=black,
	pdfstartview=Fit,
	breaklinks=true}

\begin{document}

\title{CRANE: Knowledge Editing for Reasoning MLLMs}

\author{Han Huang, Hao Wang, Mengqi Zhang, Shu Wu, Qiang Liu, Liang Wang
\thanks{Han Huang, Shu Wu, Qiang Liu and Liang Wang are with University of Chinese Academy of Sciences, Beijing 101408, China, and also with the New Laboratory of Pattern Recognition (NLPR), CASIA, Beijing 100045, China. Email: \nolinkurl{han.huang@cripac.ia.ac.cn}; \nolinkurl{qiang.liu@nlpr.ia.ac.cn}; \nolinkurl{shu.wu@nlpr.ia.ac.cn}; \nolinkurl{wangliang@nlpr.ia.ac.cn}.}
\thanks{Hao Wang is with Harbin Institute of Technology, Weihai 264209, China. Email: \nolinkurl{2023210984@stu.hit.edu.cn}.}
\thanks{Mengqi Zhang is with the School of Computer Science and Technology, Shandong University, Qingdao 266237, China. Email: \nolinkurl{mengqi.zhang@sdu.edu.cn}.}
}

\maketitle

%%%%%%%%%%%%%%%%%%%%%%%%%%%%%%%%%%%%%%%%%%%%%%%%%%%%%%%%%%%%%%%%%%%%%%%%%%%%%%%%%
\begin{abstract}
%%%%%%%%%%%%%%%%%%%%%%%%%%%%%%%%%%%%%%%%%%%%%%%%%%%%%%%%%%%%%%%%%%%%%%%%%%%%%%%%%
The emergence of reasoning multimodal large language models (MLLMs), which generate explicit chain-of-thought (CoT) reasoning within \texttt{<think>} tags before producing answers, has introduced a new challenge for knowledge editing: methods that appear successful under traditional metrics (teacher-forcing accuracy up to 100\%) can fail severely when the model's reasoning process is examined (Grounded Success as low as 0\%). We identify three failure modes: (1) \textit{Structural Collapse}, where weight-modifying methods destroy the CoT format; (2) \textit{Cognitive Dissonance}, where the model's reasoning chain actively rejects the injected edit fact based on visual evidence; and (3) \textit{Shallow Internalization}, where methods succeed on exact queries but fail on rephrase or multi-hop variants. On reasoning MLLMs, these modes interact: methods that generalize (FT, LoRA) trigger format collapse, while methods without deep modification cannot generalize. To expose these failures, we propose a CoT-aware evaluation protocol and construct \textbf{ReasonEdit-Bench}, an evaluation suite with conflict stratification, multi-level probes, and multi-hop portability tests.

Building on this diagnostic foundation, we propose \textbf{CRANE}, a retrieval-augmented framework that requires no per-edit parameter modification. CRANE combines a modality-aware dual-library retrieval system with a two-phase training strategy: Supervised Fine-Tuning (SFT) for structural initialization, followed by GRPO with a Cognitive Routing Reward that trains the model to arbitrate between visual priors and injected edit facts. On ReasonEdit-Bench, CRANE achieves 96.9\% Grounded Success on conflict scenarios and 96.9\% intermediate entity usage in multi-hop chains, while maintaining 97.6\% text-locality and 68.1\% image-locality Edit Independence. On the out-of-distribution MMEVOKE benchmark, CRANE reaches 87.0\% under gold retrieval.
\end{abstract}


%%%%%%%%%%%%%%%%%%%%%%%%%%%%%%%%%%%%%%%%%%%%%%%%%%%%%%%%%%%%%%%%%%%%%%%%%%%%%%%%%
\section{Introduction}
\label{sec:intro}
%%%%%%%%%%%%%%%%%%%%%%%%%%%%%%%%%%%%%%%%%%%%%%%%%%%%%%%%%%%%%%%%%%%%%%%%%%%%%%%%%

\IEEEPARstart{K}{nowledge} editing~\cite{meng2022rome,mitchell2022fast}, the task of updating specific facts in a deployed model while preserving unrelated knowledge, has been studied extensively on standard language and multimodal models. However, a new generation of models, exemplified by OpenAI o1~\cite{openai2024o1} and DeepSeek-R1~\cite{deepseek2025r1}, now generates explicit chain-of-thought (CoT) reasoning traces before producing answers, achieving impressive results on mathematical olympiad problems and competitive programming. This architectural shift introduces a fundamental tension: existing editing methods were designed for models that directly output answers, not for models that generate an explicit reasoning chain before answering. We find two compounding problems: (1) existing editing methods fail in previously undetected ways on these reasoning models, and (2) the standard evaluation metrics (teacher-forcing accuracy, exact match) cannot detect these failures because they never inspect the model's reasoning process.

\begin{figure}[t]
\centering
\includegraphics[width=\linewidth]{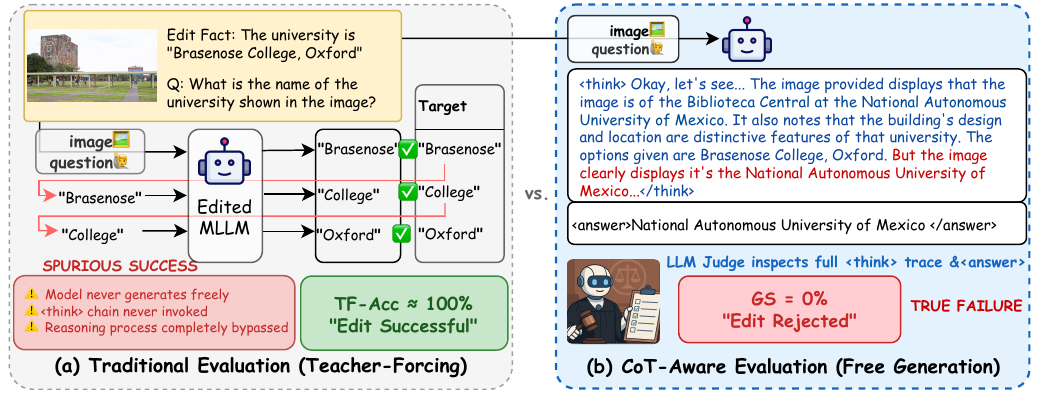}
\caption{The Spurious Success problem. \textbf{(a)} Teacher-forcing feeds target tokens as context; the model never generates freely and its \texttt{<think>} chain is never invoked, yielding TF-Acc $\approx$ 100\%. \textbf{(b)} Under free generation, the \texttt{<think>} trace explicitly rejects the edit fact based on visual evidence. Our LLM Judge inspects the full chain and reports GS = 0\%.}
\label{fig:teaser}
\end{figure}

Through a systematic study across editing methods and model families (Sections~\ref{sec:evaluation}--\ref{sec:empirical}), we find that existing methods fail on reasoning MLLMs in three distinct ways. First, \textbf{Structural Collapse}: because reasoning MLLMs depend on the precise \texttt{<think>...</think><answer>...</answer>} schema (hereafter ``CoT schema''), the weight perturbations introduced by parameter-modifying methods (FT, LoRA~\cite{hu2022lora}) and the routing rewrites of parameter-isolation methods (WISE~\cite{wang2024wise}, GRACE~\cite{grace2023}) destroy the model's ability to produce valid reasoning traces, with format rates dropping to near zero. Second, \textbf{Cognitive Dissonance}: when the edit fact contradicts the image, the model's \texttt{<think>} chain ``sees'' the visual evidence and actively rejects the injected fact, refusing to comply even when the edit is correctly stored (Fig.~\ref{fig:teaser}). Third, \textbf{Shallow Internalization}: methods that store edits in an external memory keyed by the input (e.g., GRACE) succeed on the original query but fail to generalize to rephrased or multi-hop variants. These modes interact to form a dilemma: the methods that modify parameters deeply enough to generalize are precisely those that trigger Structural Collapse, while methods that avoid deep modification cannot generalize beyond the exact query.

These failures have gone undetected because the standard evaluation protocol never inspects the model's reasoning process. Multimodal knowledge editing is conventionally measured by exact-match (EM) against the target string or teacher-forcing accuracy (TF-Acc), which feeds the target tokens as forced context~\cite{yang2025mirage}; under teacher forcing the \texttt{<think>} chain is never invoked, so the metric reports near-perfect scores even when the model would reject the edit under free generation (Fig.~\ref{fig:teaser}). To expose these failures, we propose a CoT-aware evaluation protocol that inspects the full reasoning chain, and construct \textbf{ReasonEdit-Bench}, an evaluation suite with conflict stratification, free-form knowledge expansion with multi-level internalization probes, and multi-hop portability tests.

To address these three failure modes, we propose \textbf{CRANE} (\textbf{C}ounterfactual \textbf{R}easoning \textbf{A}rbitration for Multimodal k\textbf{N}owledge \textbf{E}diting), a retrieval-augmented editing framework that requires no per-edit parameter modification, avoiding Structural Collapse by construction. CRANE pairs a modality-aware dual-library retrieval system with a two-phase training strategy: supervised fine-tuning (SFT) for structural initialization, followed by GRPO with a \textit{Cognitive Routing Reward} that teaches the model to arbitrate between visual priors and injected edit facts, directly targeting Cognitive Dissonance. Because the editing behavior is trained into the model rather than stored as a lookup, CRANE generalizes to rephrased and multi-hop queries, mitigating Shallow Internalization, while largely preserving locality on unrelated queries.

In summary, our contributions are:
\begin{enumerate}
    \item A \textbf{diagnostic framework} that identifies three failure modes of knowledge editing on reasoning MLLMs, together with a CoT-aware evaluation protocol of four metrics (Grounded Success, Edit Independence, Schema Integrity, and Intermediate Entity Used) backed by an LLM-as-a-Judge that inspects the full reasoning chain (Section~\ref{sec:evaluation}).

    \item \textbf{ReasonEdit-Bench}, an evaluation suite derived from VLKEB~\cite{vlkeb} with conflict stratification, free-form knowledge expansion with multi-level internalization probes, and multi-hop portability tests (Section~\ref{sec:empirical}).

    \item \textbf{CRANE}, a retrieval-augmented framework combining modality-aware dual-library retrieval with a contrastive projection head, SFT-based structural initialization, and GRPO with a Cognitive Routing Reward over three sub-rewards: Fact Alignment ($R_\text{align}$), Visual Suppression ($R_\text{override}$), and Epistemic Isolation ($R_\text{isolate}$) (Section~\ref{sec:method}).

    \item \textbf{Comprehensive experiments} on ReasonEdit-Bench and out-of-distribution MMEVOKE~\cite{mmevoke}, showing consistent gains over existing baselines across conflict, locality, and multi-hop scenarios (Section~\ref{sec:experiments}).
\end{enumerate}

%%%%%%%%%%%%%%%%%%%%%%%%%%%%%%%%%%%%%%%%%%%%%%%%%%%%%%%%%%%%%%%%%%%%%%%%%%%%%%%%%
\section{Related Work}
\label{sec:related}
%%%%%%%%%%%%%%%%%%%%%%%%%%%%%%%%%%%%%%%%%%%%%%%%%%%%%%%%%%%%%%%%%%%%%%%%%%%%%%%%%

\subsection{Knowledge Editing in LLMs and MLLMs}

Knowledge editing methods can be grouped into three families. \textit{Parameter-modification} methods directly update model weights: ROME~\cite{meng2022rome} locates and edits factual associations by computing a rank-one update to a specific MLP layer; MEMIT~\cite{meng2023memit} extends this to batch editing by distributing updates across multiple layers; LoRA-based fine-tuning~\cite{hu2022lora} adapts the model to new facts via low-rank weight updates. \textit{Parameter-isolation} methods preserve original weights by routing edited knowledge through auxiliary structures: GRACE~\cite{grace2023} maintains an external codebook and intercepts queries via nearest-neighbor lookup at inference time; WISE~\cite{wang2024wise} employs a side-network that selectively routes queries to either the original or the edited pathway. \textit{In-context editing} methods such as IKE~\cite{zheng2023ike} bypass weight modification entirely by prepending retrieved edit demonstrations to the prompt at inference time. In the multimodal domain, \cite{cheng2023edit,vlkeb} extend these paradigms to image-question pairs, while LiveEdit~\cite{liveedit} addresses lifelong editing via low-rank mixture-of-experts.

All of the above methods were developed and evaluated on standard MLLMs without explicit CoT reasoning chains. To our knowledge, their failure modes on reasoning MLLMs have not been systematically studied prior to this work.

\subsection{Reasoning Models and Chain-of-Thought}

The success of DeepSeek-R1~\cite{deepseek2025r1} has catalyzed widespread adoption of process-reward training and explicit CoT generation. GRPO~\cite{shao2024deepseekmath} extends PPO~\cite{schulman2017ppo} to group-relative advantage estimation, enabling stable RL training without a separate critic model. Subsequent work has explored applying RL-based CoT training to multimodal settings~\cite{liu2025visualr1,r1v}, yielding models such as Vision-R1~\cite{huang2025visionr1} that exhibit strong visual reasoning. CoT-based approaches have also been applied to specific multimodal tasks such as facial expression recognition~\cite{lan2025expllm} and visual commonsense reasoning~\cite{li2026visual}. While recent work has explored SFT+RL pipelines for text-only knowledge editing~\cite{fu2026learning}, to our knowledge no prior study has systematically analyzed how existing editing methods fail on multimodal reasoning MLLMs or proposed editing approaches specifically designed for models with explicit CoT traces.

\subsection{Visual Priors and Knowledge Conflict}

A growing body of literature documents that MLLMs exhibit strong visual anchoring: the model's output is heavily biased toward the semantic content of the input image, even when the image contradicts the correct answer~\cite{rohrbach2018object,li2023evaluating}. This visual prior creates a direct conflict with knowledge editing in conflict scenarios, where the edit fact deliberately contradicts the image content. Recent work has proposed contrastive decoding~\cite{qiang2026mitigating} and RLHF-based methods~\cite{yu2024rlhf} to reduce hallucination, but they do not address the specific problem of making a model robustly override visual priors when explicitly instructed to do so through an edit fact.

%%%%%%%%%%%%%%%%%%%%%%%%%%%%%%%%%%%%%%%%%%%%%%%%%%%%%%%%%%%%%%%%%%%%%%%%%%%%%%%%%
\section{Rethinking Evaluation: A CoT-Aware Protocol}
\label{sec:evaluation}
%%%%%%%%%%%%%%%%%%%%%%%%%%%%%%%%%%%%%%%%%%%%%%%%%%%%%%%%%%%%%%%%%%%%%%%%%%%%%%%%%

\subsection{The Illusion of Success: A Case Study}

Fig.~\ref{fig:teaser} illustrates the core problem with a real example from the VLKEB conflict split. A reasoning MLLM is shown an image of the UNAM Central Library (a visually distinctive building) along with an edit fact stating the institution is ``Brasenose College, Oxford''. Under teacher-forcing (Fig.~\ref{fig:teaser}a), the target tokens are fed as context and the model achieves near-perfect prediction accuracy without ever generating freely or invoking its \texttt{<think>} chain. Under free generation (Fig.~\ref{fig:teaser}b), the model's reasoning trace correctly identifies the building as UNAM and explicitly rejects the edit fact, outputting the visually anchored answer. This disconnect, where teacher-forcing reports success while the model's actual reasoning actively contradicts the edit, is what we term \textit{Spurious Success}.

\subsection{CoT-Aware Evaluation Metrics}
\label{subsec:metrics}

We address the Spurious Success problem with a multi-level evaluation protocol combining automated format checking with an LLM-as-a-Judge system (full templates in Appendix~A).

\textbf{Schema Integrity (Format Rate).} A rule-based check: the fraction of outputs maintaining valid CoT schema. No LLM judge is needed for this metric.

The remaining three metrics are assessed via LLM Judge prompts (P1--P3):

\textbf{Grounded Success (GS).} Our primary reliability metric (Judge P1). GS equals 1 only when (a) the surface answer matches the edit target \textit{and} (b) the judge confirms that the \texttt{<think>} trace explicitly grounds the answer in the edit fact. A model that arrives at the correct answer without referencing the edit fact scores 0 on GS but 1 on EM, precisely catching Spurious Success.

\textbf{Edit Independence (EI).} Our primary locality metric (Judge P2). EI equals 1 when neither the \texttt{<think>} trace nor the \texttt{<answer>} is contaminated by an irrelevant edit fact. A model with high EI has learned principled epistemic isolation, distinguishing when an edit fact applies and when it does not.

\textbf{Intermediate Entity Used (IU) / Used Visual Instead (UVI).} Multi-hop metrics (Judge P3). IU measures whether the edited entity is correctly propagated through a reasoning chain as an intermediate step. UVI measures regression to the visual prior instead of the edit fact. Together they capture \textit{internalization depth} beyond surface recall.

\subsection{Protocol Validation: Quantifying the Gap}

To validate that the proposed GS metric captures failures invisible to traditional TF-Acc, we compare both metrics across six editing methods on a standard MLLM (LLaVA-1.5) and a reasoning MLLM (Vision-R1). We further stratify the evaluation into \textit{normal} scenarios (image compatible with edit) and \textit{conflict} scenarios (image contradicts edit, as in Fig.~\ref{fig:teaser}), since the visual-textual relationship fundamentally affects how the model's reasoning chain engages with the edit fact. The formal construction of this stratification is detailed in Section~\ref{sec:empirical}.

\begin{figure}[htbp]
\centering
\includegraphics[width=\linewidth]{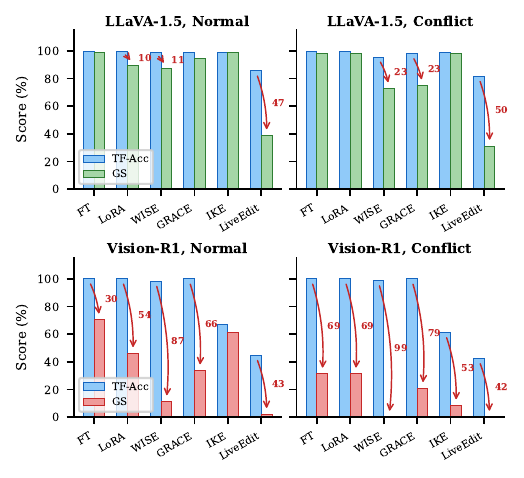}
\caption{The Metric Gap between TF-Acc and GS on the VLKEB eval set (n=200, single edit). Red arrows indicate the gap (pp). Top: standard MLLM (LLaVA-1.5). Bottom: reasoning MLLM (Vision-R1).}
\label{fig:metric_gap}
\end{figure}

Fig.~\ref{fig:metric_gap} reveals a systematic discrepancy: TF-Acc reports near-perfect scores across all methods, while GS exposes substantial failures that only our LLM Judge can detect by inspecting the reasoning chain. On reasoning MLLMs, the TF-Acc$\to$GS gap ranges from 5.5pp (IKE, normal) to 99pp (WISE, conflict). For standard MLLMs, the gap is substantially smaller (under 23pp), because there is no explicit reasoning chain that can articulate a rejection or lose its structural format. This confirms that TF-Acc is unreliable for evaluating editing on reasoning MLLMs, and that conflict settings amplify the discrepancy. \textbf{Evaluating these models requires inspecting the full CoT trace, not just the final answer or per-token predictions.}

%%%%%%%%%%%%%%%%%%%%%%%%%%%%%%%%%%%%%%%%%%%%%%%%%%%%%%%%%%%%%%%%%%%%%%%%%%%%%%%
\section{Empirical Analysis: Failure Modes of Knowledge Editing on Reasoning MLLMs}
\label{sec:empirical}
%%%%%%%%%%%%%%%%%%%%%%%%%%%%%%%%%%%%%%%%%%%%%%%%%%%%%%%%%%%%%%%%%%%%%%%%%%%%%%%

Having established the evaluation protocol and quantified the metric gap, we now conduct a systematic empirical study to characterize \textit{how} and \textit{why} existing editing methods fail on reasoning MLLMs. We first construct ReasonEdit-Bench (Section~\ref{subsec:bench}), then document three failure modes in detail. These modes differ in their relationship to reasoning capability: \textbf{Structural Collapse} is \textit{unique} to reasoning MLLMs, since standard MLLMs have no CoT format to destroy. \textbf{Cognitive Dissonance} is \textit{amplified} by reasoning MLLMs, which actively reason about contradictions rather than passively accepting them. \textbf{Shallow Internalization} is a method-level deficiency present across all model types, but on reasoning MLLMs it creates an additional dilemma: methods that modify parameters deeply enough to generalize (FT, LoRA) simultaneously trigger Structural Collapse, while methods without deep modification (GRACE, IKE) cannot generalize.

\begin{figure*}[t]
\centering
\includegraphics[width=\textwidth]{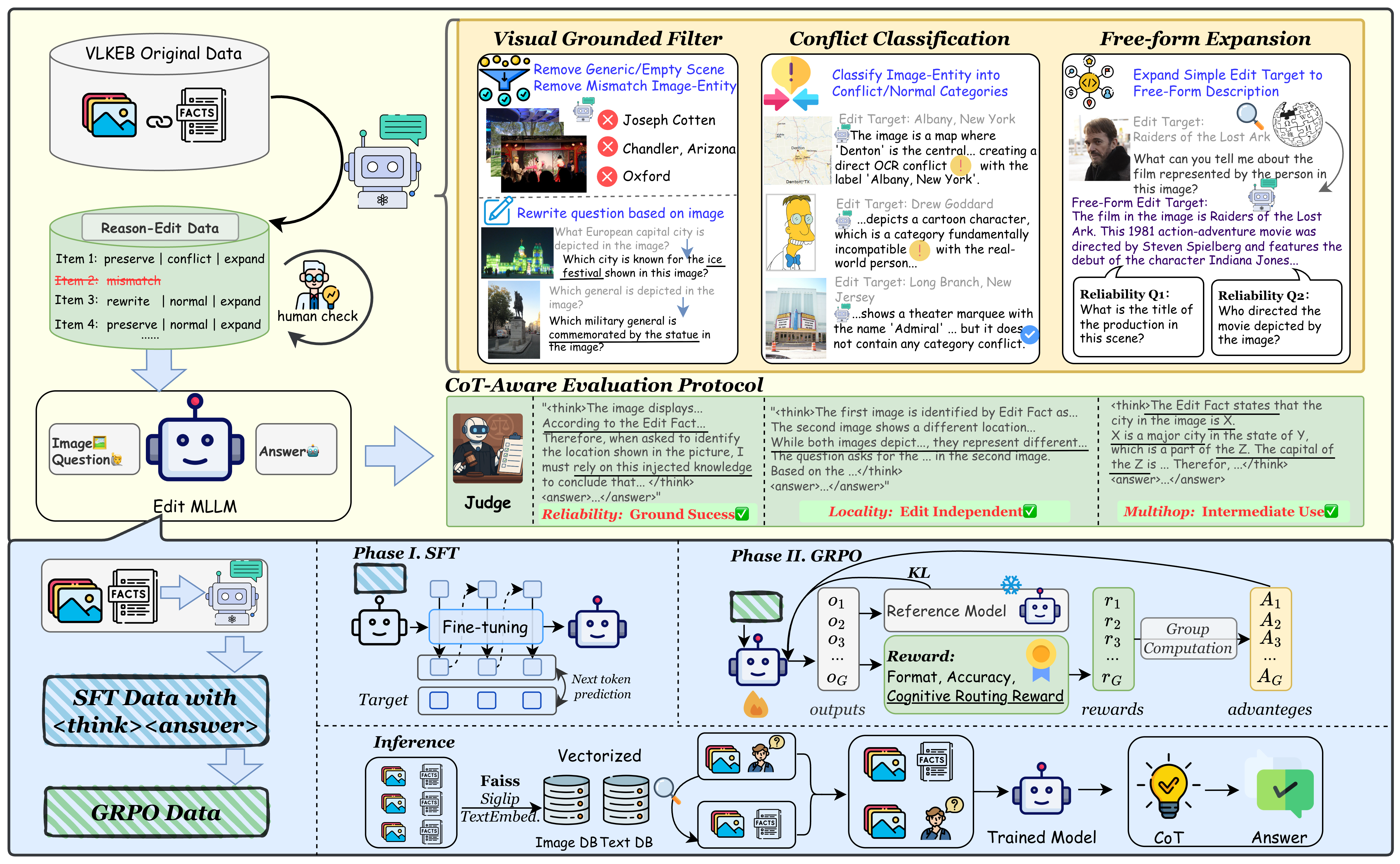}
\caption{Overview of ReasonEdit-Bench and CRANE. Top: three pipelines filter, stratify, and expand raw VLKEB data; the CoT-aware evaluation protocol then inspects full reasoning traces via LLM Judge. Bottom: CRANE combines dual-library retrieval with SFT + GRPO training.}
\label{fig:pipeline_framework}
\end{figure*}

\subsection{Constructing ReasonEdit-Bench}
\label{subsec:bench}

VLKEB~\cite{vlkeb} is a large-scale multimodal knowledge editing benchmark that pairs visual entities with factual edits. During its construction, questions and answers were generated from text triples via a language model without access to the image content, which can introduce two classes of incidental noise: (1) \textit{image-question mismatches}, where a question carries assumptions that do not hold for the paired image (e.g., asking ``which action film is shown?'' when the image depicts a comedy poster); and (2) \textit{flat conflict treatment}, where samples with strong visual contradictions (e.g., a prominently labeled movie poster) and those with weak ones are evaluated identically. To improve evaluation reliability, we apply three independent processing pipelines using Gemini-2.5-Flash~\cite{team2023gemini} as an automated annotator (see Fig.~\ref{fig:pipeline_framework}, top). In each pipeline, Gemini receives the image together with structured prompts and returns judgments or rewrites; all outputs are post-processed with rule-based validation.

\textbf{Pipeline 1 -- Visual Grounded Filter.}
Each sample's image and associated metadata are sent to Gemini for visual relevance verification. The model assesses whether the image depicts an identifiable entity matching the claimed subject. A sample is rejected if (a) the image is a generic scene with no identifiable subject, (b) the image contradicts the claimed entity at a categorical level, or (c) the visual relevance score falls below a threshold. Questions that carry wrong assumptions or awkward phrasing are rewritten by Gemini to neutral, accurate forms while preserving the answerable intent. Multi-hop portability questions are judged hop-by-hop; hops with factual errors, over-stretched reasoning, or ambiguous answers are pruned individually.

\textbf{Pipeline 2 -- Conflict Severity Classification.}
Each sample is classified as \textit{normal} or \textit{conflict} along two axes via Gemini. An \textbf{OCR conflict} occurs when the image contains text that directly names the original entity (e.g., a movie poster reading ``Apollo 13'' when the edit target is ``Primary Colors''). A \textbf{category conflict} occurs when the image category is incompatible with the edit target (e.g., an animated cartoon when the target is a real person). Samples satisfying either condition are labeled \textit{conflict}; the remainder are \textit{normal}.

\textbf{Pipeline 3 -- Free-form Knowledge Expansion.}
The original VLKEB format uses a bare entity name as the edit target, which tests only surface-level name recall. We use Gemini to expand each edit target into a two-sentence description: the first sentence anchors the entity identity, and the second adds at least one verifiable attribute fact (location, date, affiliation, etc.). A second-level probe question is then derived from the attribute sentence under the constraint that it must not name the entity, requiring the model to reason rather than recall directly. This yields two evaluation levels: \textit{Rel-1} (entity identification) and \textit{Rel-2} (attribute reasoning).

We conduct human verification on a random 5\% sample of each pipeline's output, finding $>$95\% agreement with Gemini's judgments across all three pipelines.

The resulting \textbf{ReasonEdit-Bench} comprises two evaluation splits. The \textit{filtered split} contains 1,623 normal and 1,070 conflict samples (plus 649 and 438 multi-hop counterparts), cleaned and stratified via Pipelines 1 and 2. The \textit{freeform split} mirrors the filtered split with expanded edit targets and two-level probes, constructed via Pipeline 3. The suite's diagnostic power comes from its joint use with the CoT-aware evaluation protocol of Section~\ref{sec:evaluation}; applied with a standard answer-only judge, it functions as a conventional multimodal editing benchmark.

\subsection{Phenomenon 1: Structural Collapse}

\textbf{Editing methods that rewrite model weights collapse the CoT schema almost entirely, and even activation-routing methods degrade it.} We apply representative weight-modifying methods (Fine-tuning, LoRA, WISE, GRACE, and LiveEdit) to a reasoning MLLM (Vision-R1) and measure two indicators: Format Rate (fraction of outputs with valid CoT schema) and Broken Rate (fraction of degenerate outputs exhibiting repetition loops or garbled text, assessed by our LLM Judge P0 prompt; see Appendix~A). The results in Table~\ref{tab:structural_collapse} reveal a consistent pattern: methods that modify model parameters or routing structures cause substantial degradation in the model's ability to produce the CoT schema.

The mechanism is clear: reasoning MLLMs are trained with strong format rewards that tightly couple the CoT schema tokens to the model's generation policy. Parameter-modification methods (FT, LoRA) that directly rewrite model weights, and parameter-isolation methods (WISE) that train substitute weights via gradient optimization, both reduce format compliance to 0\% by replacing layer activations with values that downstream layers have never encountered during format-reward training. Even the least disruptive methods, GRACE's codebook routing and LiveEdit's MoE adapters, degrade format compliance by altering the model's internal hidden states at inference time.

The severity is quantified in Table~\ref{tab:structural_collapse}: on Vision-R1, FT/LoRA/WISE all achieve 0\% Format Rate (with WISE additionally producing 68\% broken outputs), GRACE retains only 0.5\%. LiveEdit (53.9\%) and IKE (85.5\%) are the least affected, with IKE fully preserving format since it modifies no weights.

\begin{table}[htbp]
\centering
\caption{Structural Collapse on Vision-R1 (Format / Broken Rate).}
\label{tab:structural_collapse}
\begin{tabular}{lcc|lcc}
\toprule
\textbf{Method} & \textbf{FR$\uparrow$} & \textbf{BR$\downarrow$} & \textbf{Method} & \textbf{FR$\uparrow$} & \textbf{BR$\downarrow$} \\
\midrule
FT     & 0.0\%  & 27.5\% & GRACE    & 0.5\%  & 7.5\% \\
LoRA   & 0.0\%  & 18.5\% & LiveEdit & 53.9\% & 1.5\% \\
WISE   & 0.0\%  & 68.0\% & IKE      & 85.5\% & 0.0\% \\
\bottomrule
\end{tabular}
\end{table}

\subsection{Phenomenon 2: Cognitive Dissonance}

\textbf{Even when the edit is correctly stored, the model's reasoning chain overrides it whenever the image disagrees---the failure lies at the reasoning level, not the storage level.} Cognitive Dissonance occurs when the model's \texttt{<think>} trace acknowledges the edit fact but ultimately sides with visual evidence, producing the pre-edit answer. While this phenomenon can occur with any editing method in conflict scenarios (even FT achieves only 22.4\% conflict Img Rep GS on Vision-R1), it is most cleanly observable with in-context editing (IKE), which preserves Schema Integrity and allows direct inspection of the reasoning chain.

When the input image visually contradicts the injected edit fact, the \texttt{<think>} trace typically follows a three-stage pattern: (1) description of the visual content using original knowledge, (2) acknowledgment of the contradiction with the injected fact, (3) rejection of the edit fact in favor of visual evidence.

As shown by the GS metrics in Table~\ref{tab:main_results}, cognitive dissonance is most severe on conflict samples: IKE on Vision-R1 achieves only 10.8\% conflict Image-Rephrase GS despite 57.2\% normal, a 46pp drop driven by visual-textual conflict. IKE on Qwen2.5-VL is even worse, dropping to 6.1\% conflict GS from 23.7\% normal. This failure is especially pronounced in multi-hop reasoning, where the model must propagate the injected fact through multiple inference steps, each providing another opportunity for the visual prior to re-assert itself, as reflected by the UVI scores reaching 98.8--100\% for IKE across all models (Fig.~\ref{fig:multihop}).

\subsection{Phenomenon 3: Shallow Internalization}

\begin{figure}[t]
\centering
\includegraphics[width=\linewidth]{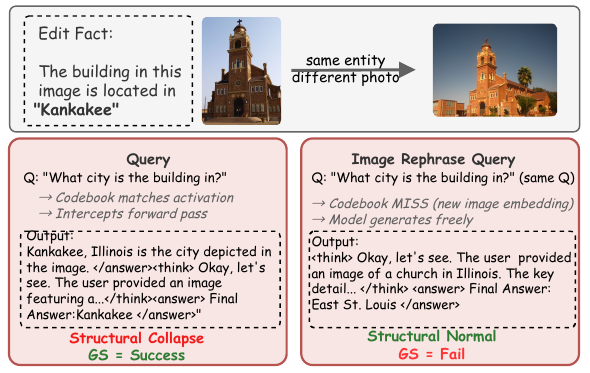}
\caption{Shallow Internalization (GRACE + Vision-R1). Same question, same entity, different photo. Left: codebook hit, correct answer but format collapse. Right: codebook miss, model falls back to original knowledge (GS = 0\%).}
\label{fig:failure_modes}
\end{figure}

\textbf{Methods that store edits in an external memory keyed by the input succeed on the original query but fail to generalize to rephrased or reformulated versions of the same knowledge.} Beyond format and reasoning-chain failures, a third deficiency concerns the \textit{depth} of knowledge internalization: a method may produce correct answers on the exact original query yet fail when the same knowledge is probed differently. Table~\ref{tab:internalization_depth} reports GS across five increasingly demanding probes on Qwen2.5-VL: (1) \textit{Rewrite}, the standard format asking directly for the edited entity; (2) \textit{Text Rephrase}, the same question rephrased textually; (3) \textit{Image Rephrase}, the same question with a different photograph of the same entity; (4) \textit{Freeform Rel\_1}, entity identification from expanded edit fact; and (5) \textit{Freeform Rel\_2}, attribute reasoning requiring genuine knowledge internalization.

The results reveal two distinct failure patterns. GRACE achieves 91.2\% Rewrite GS but collapses to 0.2\%/0.3\% on Text/Image Rephrase, the most extreme evidence of Shallow Internalization. As illustrated in Fig.~\ref{fig:failure_modes}, GRACE's codebook intercepts only exact-match queries; a different photo of the same entity bypasses it entirely, causing the model to fall back to its original parametric knowledge. IKE presents a different pattern: its Rewrite GS is only 21.8\% on Qwen2.5-VL, far below LLaVA's 97--100\% (Table~\ref{tab:main_results}), because Qwen2.5-VL does not blindly follow the injected edit fact but applies its own visual and parametric knowledge to override it. Note that in this evaluation the edit fact is directly injected (no retrieval involved), so the low GS reflects Cognitive Dissonance rather than retrieval failure. Text Rephrase (21.3\%) and Image Rephrase (23.7\%) remain at similar levels, confirming that the bottleneck is the model's refusal to comply, not the query format. IKE's anomalously high Freeform Rel\_2 (89.3\%) does not indicate deep internalization: in freeform mode, the edit fact is injected as a long descriptive text containing attribute information, so Rel\_2 (attribute query) is directly answerable by copying from the in-context text without genuine reasoning. FT and LoRA maintain stable GS across all levels (78--97\%), with minor fluctuations within sampling noise (n=100 per split), demonstrating genuine parametric knowledge modification. However, as shown in Table~\ref{tab:structural_collapse}, this comes at the cost of severe Structural Collapse on reasoning MLLMs.

\begin{table}[htbp]
\centering
\caption{Shallow Internalization: GS across increasingly demanding probes (Qwen2.5-VL, normal). Txt/Img Rep = text/image rephrase. FF Rel\_1/2 = freeform entity/attribute probe. Drop = Rewrite $\to$ Rel\_1.}
\label{tab:internalization_depth}
\resizebox{\linewidth}{!}{
\begin{tabular}{lcccccc}
\toprule
\textbf{Method}
  & \textbf{Rewrite}
  & \textbf{Txt Rep}
  & \textbf{Img Rep}
  & \textbf{FF Rel\_1}
  & \textbf{FF Rel\_2}
  & \textbf{Rew$\to$Rel\_1} \\
\midrule
FT       & 78.4\% & 80.6\% & 79.8\% & 72.6\% & 83.7\% & $-$6pp \\
LoRA     & 93.8\% & 96.7\% & 92.4\% & 90.8\% & 81.4\% & $-$3pp \\
WISE     & 46.3\% & 37.8\% & 49.7\% & 38.7\% & 34.2\% & $-$8pp \\
GRACE    & 91.2\% & \textbf{0.2\%} & \textbf{0.3\%} & \textbf{0.2\%} & 4.8\% & \textbf{$-$91pp} \\
IKE      & 21.8\% & 21.3\% & 23.7\% & 54.6\% & 89.3\% & +33pp \\
LiveEdit & 28.7\% & 26.8\% & 22.4\% & 5.7\%  & 7.2\%  & $-$23pp \\
\bottomrule
\end{tabular}
}
\end{table}

%%%%%%%%%%%%%%%%%%%%%%%%%%%%%%%%%%%%%%%%%%%%%%%%%%%%%%%%%%%%%%%%%%%%%%%%%%%%%%%
\section{CRANE Framework}
\label{sec:method}
%%%%%%%%%%%%%%%%%%%%%%%%%%%%%%%%%%%%%%%%%%%%%%%%%%%%%%%%%%%%%%%%%%%%%%%%%%%%%%%

The three failure modes documented in Section~\ref{sec:empirical} motivate a different approach to knowledge editing for reasoning MLLMs. As shown in Fig.~\ref{fig:pipeline_framework}(b), CRANE proceeds from three principles: (1) \textit{retrieve accurately}, the correct edit fact must be identified at inference time without any weight modification; (2) \textit{format first}, the model must master the CoT schema before learning to reason about edit facts; and (3) \textit{resolve the conflict}, the model must be explicitly trained to arbitrate between visual priors and textual edit facts through a reinforcement signal.

\subsection{Modality-Aware Retrieval System}

Cross-modal retrieval methods~\cite{pu2025deep} typically learn joint embeddings, but in our conflict scenarios the query image and the edit fact point in opposing semantic directions. To avoid this interference, we adopt a dual-library architecture with separate indices for each modality.

\textbf{Dual-library with modality-conditional routing.} We maintain two independent FAISS indices~\cite{johnson2019faiss}: an \textit{image index} containing SigLIP~\cite{zhai2023sigmoid} embeddings of edit-fact images, and a \textit{text index} containing SigLIP text embeddings of edit-fact descriptions. Routing is deterministic and conditioned solely on query modality: if the query contains an image (normal, conflict, or locality scenarios), the query image is encoded and searched against the image index; if the query is text-only (text\_locality), the question text is encoded and searched against the text index. This design keeps each modality's retrieval path independent.

\textbf{Contrastive projection head.} While SigLIP provides strong general-purpose visual features, we observe that different entities often lie close together in the 1152-dimensional SigLIP space (e.g., two actors' headshots, two city skylines), causing retrieval errors. We train a lightweight linear projection head $f: \mathbb{R}^{1152} \to \mathbb{R}^{256}$ using symmetric InfoNCE loss~\cite{oord2018representation}:
\begin{equation}
\mathcal{L} = -\frac{1}{2N}\sum_{i=1}^{N}\left[\log\frac{e^{z_{a_i}\cdot z_{b_i}/\tau}}{\sum_j e^{z_{a_i}\cdot z_{b_j}/\tau}} + \log\frac{e^{z_{b_i}\cdot z_{a_i}/\tau}}{\sum_j e^{z_{b_i}\cdot z_{a_j}/\tau}}\right]
\end{equation}
where $z_{a_i} = f(\text{SigLIP}(x_{a_i}))$ and $z_{b_i} = f(\text{SigLIP}(x_{b_i}))$ are L2-normalized projected embeddings of the $i$-th positive pair: $x_{a_i}$ is an edit-fact image and $x_{b_i}$ is a rephrase image of the same entity (different photograph), $\tau = 0.07$ is the temperature, and $N$ is the batch size. To prevent data leakage, we exclude all entities appearing in the evaluation split, yielding 3,579 training pairs across 2,792 unique entities. This projection improves hit@1 from 69.7\% to 90.4\% on unseen entities (+20.7pp).

\textbf{Confidence-threshold defense.} In locality scenarios, the query image depicts a \textit{different} entity than any edit fact in the database, so all retrievals are semantically irrelevant. However, proximity in feature space occasionally returns an edit fact for the wrong entity with non-trivial similarity. We observe that legitimate reliability-scenario hits score significantly higher than spurious locality hits (mean: 0.817 vs.\ 0.595). Setting a confidence threshold $\theta = 0.55$ filters 33.3\% of spurious locality hits while retaining 98.7\% of correct reliability retrievals. Queries scoring below $\theta$ receive a fallback ``no edit fact'' prompt, instructing the model to answer from world knowledge.

\subsection{Phase I: Supervised Fine-Tuning for Initialization}

Before the model can learn to reason about edit facts, it must master the structural contract of CoT editing: always produce the CoT schema, always cite relevant edit facts within \texttt{<think>}, and recognize when edit facts are irrelevant (locality scenarios). We accomplish this through a supervised fine-tuning phase on 6,000 high-quality CoT examples generated by Gemini Flash.

\textbf{Data construction.} We generate SFT data using Gemini Flash as an annotator. For each sample, we provide the edit fact, the query image, and the question, then prompt Gemini to produce a structured reasoning response covering the appropriate behavior for that scenario (e.g., citing the edit fact for normal, acknowledging conflict then overriding for conflict, recognizing irrelevance for locality). We generate samples covering all four editing scenarios and validate each output for correctness and format compliance.

\textbf{Training configuration.} We fine-tune Qwen2.5-VL-7B-Instruct for 2 epochs on 6,000 balanced samples (2,000 normal, 2,000 conflict, 1,000 locality, 1,000 text\_locality). We freeze the Visual Transformer (VIT) and the vision-language aligner, training only the language model component. This prevents catastrophic forgetting of visual representations while establishing the desired CoT reasoning patterns.

\textbf{Role of SFT.} Critically, SFT is \textit{not} intended to inject factual knowledge. The edit facts seen during SFT are not the same as those used at evaluation time. Rather, SFT teaches the model \textit{meta-skills}: how to recognize and cite an edit fact, how to acknowledge and override a visual-textual conflict, and how to identify irrelevance. These meta-skills generalize across all edit facts encountered at inference time.

\subsection{Phase II: Counterfactual Arbitration via GRPO}

Building on the SFT-initialized model, we apply Group Relative Policy Optimization (GRPO)~\cite{shao2024deepseekmath} to refine the model's behavior through outcome-based reward signals. GRPO generates $G=8$ candidate responses per prompt, computes group-relative advantages, and optimizes with clipping ($\epsilon=0.2$) and KL regularization ($\beta=0.05$). The total reward combines three equally-weighted components:
\begin{equation}
R_\text{total} = R_\text{accuracy} + R_\text{format} + R_\text{reasoning}
\end{equation}

\textbf{Accuracy Reward} ($R_\text{accuracy}$) uses soft matching: $\max(\text{F1}_\text{token}(p, g),\, \mathbf{1}[g \subseteq \text{normalize}(p)])$, where $p$ is the predicted answer and $g$ the ground truth.

\textbf{Format Reward} ($R_\text{format}$) is binary: 1 if output contains valid CoT schema, 0 otherwise.

\textbf{Cognitive Routing Reward} ($R_\text{reasoning}$) is the key novelty in CRANE's reward design. Each training sample carries a \texttt{sample\_type} label (normal, conflict, locality, or text\_locality), and $R_\text{reasoning}$ applies a \textit{different} sub-reward depending on this label. This teaches the model to adopt distinct reasoning strategies for different scenarios:

\begin{itemize}
\item \textbf{Normal samples} $\to$ $R_\text{align}$: rewards explicit citation of the edit fact in \texttt{<think>}, encouraging the model to ground its answer in the provided knowledge.
\item \textbf{Conflict samples} $\to$ $R_\text{override}$: rewards the three-stage pattern of acknowledging visual evidence, applying a pivot word (\textit{however, but, despite}), and committing to the edit fact. Full reward (1.0) requires both pivot and edit-fact reference; partial (0.5) for one; zero otherwise.
\item \textbf{Locality / text\_locality samples} $\to$ $R_\text{isolate}$: for image-locality, rewards recognizing the query image as distinct from the edit-fact image; for text-locality, penalizes hallucinating a non-existent image and rewards explicit recognition of irrelevance.
\end{itemize}

The complete reward formulations with keyword lists are provided in Appendix~B.

%%%%%%%%%%%%%%%%%%%%%%%%%%%%%%%%%%%%%%%%%%%%%%%%%%%%%%%%%%%%%%%%%%%%%%%%%%%%%%%
\section{Experiments and Analysis}
\label{sec:experiments}
%%%%%%%%%%%%%%%%%%%%%%%%%%%%%%%%%%%%%%%%%%%%%%%%%%%%%%%%%%%%%%%%%%%%%%%%%%%%%%%

\subsection{Experimental Setup}

\textbf{Base model.} We use Qwen2.5-VL-7B-Instruct~\cite{wang2024qwen2vl} (7B parameters) as the foundation model for CRANE. Qwen2.5-VL is a strong general-purpose MLLM without built-in CoT reasoning; CRANE's SFT+GRPO training equips it with structured reasoning capability specifically for knowledge editing. We choose not to build on Vision-R1 because its pre-existing CoT policy conflicts with editing-specific training (see Structural Collapse results). Vision-R1-7B~\cite{huang2025visionr1} (7B) instead serves as an evaluation-only reasoning MLLM baseline, and LLaVA-1.5-7B~\cite{liu2024improved} (7B) as the non-reasoning MLLM baseline.

\textbf{Baselines.} Parameter-editing baselines include Fine-tuning (FT), LoRA, WISE, GRACE, and LiveEdit, all evaluated in single-edit mode (model reset between edits) with the edit fact directly provided. In-context editing baseline IKE is applied to Qwen2.5-VL, Vision-R1, and LLaVA-1.5, also with the edit fact directly injected (no retrieval involved). CRANE uses its own dual-library retrieval system at inference time.

\textbf{Evaluation.} All models are evaluated on ReasonEdit-Bench under free autoregressive generation using the CoT-aware protocol of Section~\ref{sec:evaluation}: Schema Integrity (Format Rate), LLM-as-a-Judge (Gemini-2.5-Flash) for Grounded Success (GS), Edit Independence (EI), Intermediate Entity Used (IU), and Used Visual Instead (UVI). Results are reported on the filtered split (1,623 normal, 1,070 conflict), freeform split, and multi-hop subsets.

\textbf{CRANE ablations.} We compare: CRANE (full system), w/o Reasoning Reward (accuracy + format only), w/o SFT Init. (GRPO from base directly), and SFT Only (no GRPO).

\subsection{Main Results}

\begin{figure}[htbp]
\centering
\includegraphics[width=\linewidth]{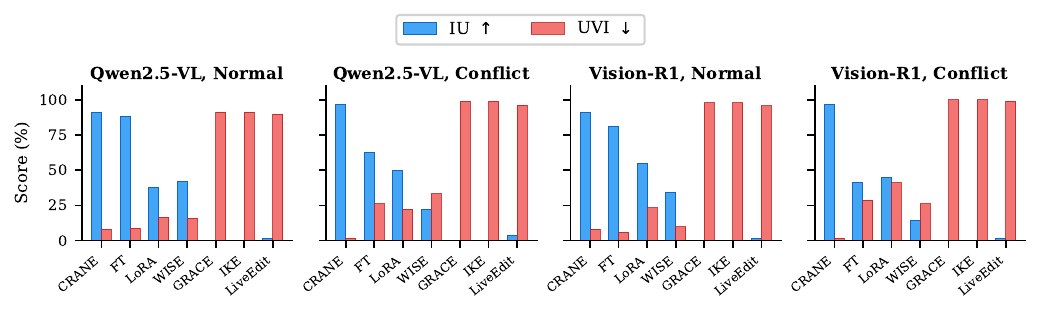}
\caption{Multi-hop reasoning: IU (edit knowledge propagated, higher is better) vs. UVI (visual prior regression, lower is better). GRACE/IKE/LiveEdit achieve near-zero IU with near-100\% UVI across all settings. CRANE maintains IU $>$ 91\% with UVI $<$ 9\%.}
\label{fig:multihop}
\end{figure}

Table~\ref{tab:main_results} and Fig.~\ref{fig:multihop} present results across reliability, locality, and multi-hop dimensions. We highlight three key findings.

\textbf{Finding 1: CRANE handles the conflict scenario where all baselines struggle.} CRANE achieves Img Rep GS = 96.9\% on conflict, higher than its own normal GS (90.5\%). This counterintuitive result arises because CRANE's $R_\text{override}$ reward explicitly trains the model to recognize and override visual contradictions; conflict scenarios activate this trained behavior reliably. In contrast, baselines show the opposite pattern: LoRA drops from 92.4\% (normal) to 86.9\% (conflict) on Qwen2.5-VL, and IKE collapses from 23.7\% to 6.1\%. The gap widens further on Vision-R1, where LoRA drops to 73.4\%/43.2\% and IKE to 57.2\%/10.8\%. On LLaVA-1.5 (no reasoning traces), IKE maintains 99.7\%/92.3\%, confirming that the degradation is specific to models capable of actively reasoning against the edit fact.

\textbf{Finding 2: High locality and high reliability are not mutually exclusive, but require different mechanisms.} CRANE achieves Text EI = 97.6\% alongside 96.9\% conflict GS. Its lower Image EI (68.1\%) reflects a specific mechanism: the retrieval system occasionally provides an irrelevant edit fact (Locality False Hit = 18.5\%, Table~\ref{tab:retrieval}), and the model sometimes engages with it. This is a retrieval-level issue, not a reasoning-level failure. Baselines that show near-perfect Image EI (GRACE 99.5\%, LiveEdit 99.5\%) achieve this because their editing mechanism is too weak to affect the model in the first place (conflict Img Rep GS $<$1\% and 6.3\% respectively). IKE on Vision-R1 shows the worst locality (51.5\% Image EI), because reasoning MLLMs' \texttt{<think>} chains are more susceptible to contamination by any injected context.

\textbf{Finding 3: Multi-hop propagation separates genuine knowledge integration from surface pattern matching.} CRANE achieves conflict IU = 96.9\% with UVI = 1.7\% (Fig.~\ref{fig:multihop}). GRACE, IKE, and LiveEdit all achieve IU $\leq$ 0.4\% with UVI $>$ 91\%, even in normal scenarios where no visual conflict exists. This means these methods never engage with the edit fact at the reasoning level: the model derives intermediate entities from visual recognition regardless of the injected knowledge. Their non-zero rewrite GS (Table~\ref{tab:internalization_depth}) is achieved through shallow matching without genuine knowledge propagation. FT achieves moderate IU (62.5\% conflict on Qwen) because parameter modification does partially internalize the entity name, but with 26\% UVI, it still frequently regresses to visual priors when multi-step reasoning is required.

\begin{table*}[htbp]
\centering
\caption{Main results on ReasonEdit-Bench (GS = Grounded Success \%, EI = Edit Independence \%). Baselines use single-edit mode; CRANE uses retrieval-augmented inference. Bold = best within the Qwen2.5-VL group (same base model as CRANE).}
\label{tab:main_results}
\begin{tabular}{ll|cc|cc|cc|cc|cc}
\toprule
\multirow{2}{*}{\textbf{Model}} & \multirow{2}{*}{\textbf{Method}}
  & \multicolumn{2}{c|}{\textbf{Txt Rep (GS $\uparrow$)}}
  & \multicolumn{2}{c|}{\textbf{Img Rep (GS $\uparrow$)}}
  & \multicolumn{2}{c|}{\textbf{FF Rel\_1 (GS $\uparrow$)}}
  & \multicolumn{2}{c|}{\textbf{FF Rel\_2 (GS $\uparrow$)}}
  & \multicolumn{2}{c}{\textbf{Locality (EI $\uparrow$)}} \\
\cmidrule{3-12}
  & & \textbf{Norm} & \textbf{Conf}
  & \textbf{Norm} & \textbf{Conf}
  & \textbf{Norm} & \textbf{Conf}
  & \textbf{Norm} & \textbf{Conf}
  & \textbf{Text} & \textbf{Img} \\
\midrule
\multirow{6}{*}{LLaVA-1.5}
& FT       & 96.3\% & 94.7\% & 98.6\% & 91.8\% & 93.4\% & 89.7\% & 70.2\% & 55.3\% & 99.2\% & 49.3\% \\
& LoRA     & 87.6\% & 88.4\% & 90.3\% & 85.1\% & 89.7\% & 91.6\% & 81.4\% & 56.2\% & 98.8\% & 27.5\% \\
& WISE     & 85.7\% & 85.9\%& 89.2\% & 75.8\% & 38.4\% & 41.6\% & 42.3\% & 28.1\% & 99.5\% & 61.4\% \\
& GRACE    & 1.2\%  & 0.3\% & 3.7\%  & 2.8\%  & 1.1\%  & 0.2\%  & 19.6\% & 7.3\%  & 100.0\%& 98.5\% \\
& IKE      & 96.8\% & 89.4\%& 99.7\% & 92.3\% & 97.6\% & 78.4\% & 100.0\%& 98.7\% & 79.9\% & 46.2\% \\
& LiveEdit & 43.8\% & 22.4\%& 40.3\% & 21.7\% & 13.2\% & 9.8\%  & 26.7\% & 15.4\% & 100.0\% & 98.8\% \\
\midrule
\multirow{6}{*}{Qwen2.5-VL}
& FT       & 80.6\% & 53.7\%& 79.8\% & 45.3\% & 72.6\% & 59.8\% & 83.7\% & 57.4\% & 84.5\% & 44.3\% \\
& LoRA     & \textbf{96.7\%} & 86.2\%& \textbf{92.4\%} & 86.9\% & \textbf{90.8\%} & 82.3\% & 81.4\% & 60.7\% & \textbf{99.5\%} & 51.8\% \\
& WISE     & 37.8\% & 16.8\%& 49.7\% & 24.6\% & 38.7\% & 17.8\% & 34.2\% & 25.7\% & 98.5\% & 55.5\% \\
& GRACE    & 0.2\%  & 0.0\% & 0.3\%  & 0.0\%  & 0.2\%  & 0.0\%  & 4.8\%  & 3.7\%  & 100.0\%& \textbf{99.5\%} \\
& IKE      & 21.3\% & 5.2\% & 23.7\% & 6.1\%  & 54.6\% & 16.4\% & 89.3\% & 76.8\% & 95.5\% & 84.5\% \\
& LiveEdit & 26.8\% & 5.8\% & 22.4\% & 6.3\%  & 5.7\%  & 0.4\%  & 7.2\%  & 8.4\%  & 99.7\% & 99.5\% \\
\midrule
\multirow{6}{*}{Vision-R1}
& FT       & 65.8\% & 31.2\%& 68.7\% & 22.4\% & 23.6\% & 9.3\%  & 54.8\% & 28.6\% & 59.5\% & 47.5\% \\
& LoRA     & 76.8\% & 54.7\%& 73.4\% & 43.2\% & 65.8\% & 51.6\% & 53.7\% & 46.8\% & 97.2\% & 44.5\% \\
& WISE     & 27.3\% & 7.6\% & 35.8\% & 7.2\%  & 38.6\% & 14.3\% & 39.7\% & 27.6\% & 86.3\% & 39.2\% \\
& GRACE    & 0.4\%  & 0.0\% & 0.2\%  & 0.0\%  & 0.3\%  & 0.0\%  & 7.8\%  & 3.6\%  & 100.0\%& 99.5\% \\
& IKE      & 52.6\% & 9.3\% & 57.2\% & 10.8\% & 72.8\% & 21.6\% & 84.3\% & 74.7\% & 55.2\% & 51.5\% \\
& LiveEdit & 2.3\%  & 0.8\% & 1.2\%  & 0.6\%  & 0.3\%  & 0.2\%  & 11.6\% & 4.7\%  & 99.8\% & 99.5\% \\
\midrule
Ours
& \textbf{CRANE} & 90.5\% & \textbf{96.9\%} & 90.5\% & \textbf{96.9\%} & 90.3\% & \textbf{96.1\%} & \textbf{91.1\%} & \textbf{97.4\%} & 97.6\% & 68.1\% \\
\bottomrule
\end{tabular}
\end{table*}

% Original multi-hop table moved to supplement for precise numbers:
% \begin{table}[htbp]
% \centering
% \caption{Multi-hop Reasoning Results on ReasonEdit-Bench.}
% \label{tab:multihop}
% ... (see appendix for full table)
% \end{table}

\subsection{Ablation Study}

Table~\ref{tab:ablation} isolates the contribution of each training component.

\textbf{SFT is the structural foundation.} Without SFT (w/o SFT Init.), the model lacks the meta-skills to engage with edit facts: conflict GS drops to 79.1\% ($-$17.8pp), multi-hop IU collapses to 35.9\% ($-$61pp), and UVI rises to 56.4\%, meaning the model reverts to visual priors in over half of multi-hop queries. Format compliance also degrades (86.2\% vs.\ 99.9\%). SFT establishes the CoT schema and teaches the basic edit-fact engagement patterns that GRPO then refines.

\textbf{GRPO improves locality without sacrificing reliability.} Comparing CRANE to SFT-Only, GRPO's primary effect is on image-locality EI: 68.1\% vs.\ 50.1\% (+18pp). Reliability and multi-hop metrics are comparable (conflict GS 96.9\% vs.\ 96.9\%, IU 96.9\% vs.\ 97.3\%). This indicates that SFT already achieves high reliability, while GRPO's reinforcement signal teaches the model \textit{when to ignore} irrelevant edit facts.

\textbf{The Reasoning Reward targets locality specifically.} Removing $R_\text{reasoning}$ (w/o Reasoning Reward) drops image EI from 68.1\% to 58.4\% ($-$9.7pp) while leaving all other metrics unchanged. The locality sub-reward ($R_\text{isolate}$) is directly responsible for this improvement, confirming that scenario-conditional rewards can address locality without reliability cost.

\begin{table}[htbp]
\centering
\caption{Ablation study of CRANE training components across all evaluation dimensions.}
\label{tab:ablation}
\resizebox{\linewidth}{!}{
\begin{tabular}{l|cc|cc|ccc|c}
\toprule
\multirow{2}{*}{\textbf{Configuration}}
  & \multicolumn{2}{c|}{\textbf{Reliability (GS)}}
  & \multicolumn{2}{c|}{\textbf{Locality (EI)}}
  & \multicolumn{3}{c|}{\textbf{Multi-hop Conflict}}
  & \textbf{Schema} \\
\cmidrule{2-9}
  & \textbf{Normal} & \textbf{Conflict}
  & \textbf{Image} & \textbf{Text}
  & \textbf{Acc} & \textbf{IU $\uparrow$} & \textbf{UVI $\downarrow$}
  & \textbf{FR $\uparrow$} \\
\midrule
\textbf{CRANE (full)}    & \textbf{90.5\%} & \textbf{96.9\%} & \textbf{68.1\%} & \textbf{97.6\%} & \textbf{40.8\%} & \textbf{96.9\%} & \textbf{1.7\%} & 99.9\% \\
w/o Reasoning Reward     & 90.9\% & 97.1\% & 58.4\% & 96.5\% & 40.0\% & 96.9\% & 1.5\% & 99.7\% \\
w/o SFT Init.           & 75.3\% & 79.1\% & 52.3\% & 75.3\% & 27.3\% & 35.9\% & 56.4\% & \textbf{86.2\%} \\
SFT Only                 & 91.2\% & 96.9\% & 50.1\% & 97.5\% & 38.3\% & 97.3\% & 1.6\% & 99.9\% \\
\bottomrule
\end{tabular}
}

\end{table}

\subsection{Retrieval System Efficacy}

Table~\ref{tab:retrieval} shows the impact of each retrieval component. The contrastive projection head is the largest contributor, improving normal hit@1 from 69.7\% (raw SigLIP) to 91.6\% (+21.9pp) by learning entity-discriminative features from same-entity image pairs. The Qwen2.5-VL ViT (57.9\%) performs worse than SigLIP because it is trained for understanding rather than retrieval.

The threshold ($\theta = 0.55$) addresses the locality problem identified in Finding 2: it reduces false edit-fact injection from 18.5\% to 12.3\%, improving image-locality EI from 68.1\% to 75.4\% (+7.3pp) at a cost of only 1.2pp in normal hit@1. This confirms that CRANE's image-locality gap is partly a retrieval issue addressable without retraining the model.

\begin{table}[htbp]
\centering
\caption{Retrieval system efficacy and downstream impact on CRANE.}
\label{tab:retrieval}
\resizebox{\linewidth}{!}{
\begin{tabular}{lccccc}
\toprule
\textbf{Strategy} & \textbf{Normal hit@1} & \textbf{Conflict hit@1} & \textbf{Loc.\ False Hit $\downarrow$} & \textbf{Normal GS $\uparrow$} & \textbf{Loc.\ EI $\uparrow$} \\
\midrule
Qwen2.5-VL ViT             & 57.9\% & 48.7\% & --- & --- & --- \\
SigLIP Base                 & 69.7\% & 93.9\% & 18.5\% & --- & --- \\
\midrule
SigLIP + Contrastive Proj.  & 91.6\% & 97.3\% & 18.5\% & 90.5\% & 68.1\% \\
\textbf{+ Threshold ($\theta$=0.55)} & 90.4\% & 97.0\% & \textbf{12.3\%} & 88.4\% & \textbf{75.4\%} \\
\bottomrule
\end{tabular}
}
\end{table}

\subsection{Out-of-Distribution Generalization on MMEVOKE}

To assess whether CRANE's reasoning ability transfers beyond VLKEB, we evaluate on MMEVOKE~\cite{mmevoke}, which covers news images and Wikipedia entities with different visual characteristics than the training distribution.

\textbf{Retrieval requires domain adaptation.} Unlike VLKEB where queries are routed to a single modality index (Section~\ref{sec:method}), MMEVOKE queries benefit from combining both image and text similarity scores, since neither modality alone works across domains: image-only achieves 70.1\% on wiki but only 19.1\% on news; text-only shows the opposite. We apply weighted fusion ($\alpha \cdot s_\text{img} + (1{-}\alpha) \cdot s_\text{text}$, with $\alpha_\text{news}$=0.30, $\alpha_\text{wiki}$=0.60) to achieve 74.5\%/79.6\%.

\textbf{CRANE's reasoning ability generalizes.} With fusion retrieval, CRANE achieves 69.8\% overall CEM vs.\ 55.2\% for the base model (+14.6pp). Under gold retrieval (oracle edit facts), CRANE reaches 87.0\% vs.\ base 67.8\% (+19.2pp). The 17pp gap between gold and fusion performance reveals that the retrieval system, not the reasoning model, is the current bottleneck for OOD deployment. Notably, CRANE with imperfect retrieval (69.8\%) already surpasses the base model with perfect retrieval (67.8\%), confirming that CRANE's trained reasoning patterns transfer to unseen entity domains.

\begin{table}[htbp]
\centering
\caption{Out-of-distribution generalization on MMEVOKE. CEM = Cover Exact Match. Base = Qwen2.5-VL-7B-Instruct. Gold = ground-truth edit fact directly provided (no retrieval).}
\label{tab:mmevoke}
\begin{tabular}{llccc}
\toprule
\textbf{Model} & \textbf{Retrieval} & \textbf{News} & \textbf{Wiki} & \textbf{Overall} \\
\midrule
Base  & Image Only     & 32.1\% & 57.3\% & 44.7\% \\
Base  & Text Only      & 41.8\% & 19.7\% & 30.8\% \\
Base  & Fusion         & 46.7\% & 63.6\% & 55.2\% \\
Base  & Gold (Oracle)  & 58.4\% & 77.2\% & 67.8\% \\
\midrule
CRANE & Fusion         & \textbf{66.8\%} & \textbf{72.7\%} & \textbf{69.8\%} \\
CRANE & Gold (Oracle)  & 83.0\% & 91.0\% & 87.0\% \\
\bottomrule
\end{tabular}
\end{table}

%%%%%%%%%%%%%%%%%%%%%%%%%%%%%%%%%%%%%%%%%%%%%%%%%%%%%%%%%%%%%%%%%%%%%%%%%%%%%%%
\section{Discussion and Limitations}
\label{sec:discussion}
%%%%%%%%%%%%%%%%%%%%%%%%%%%%%%%%%%%%%%%%%%%%%%%%%%%%%%%%%%%%%%%%%%%%%%%%%%%%%%%

\textbf{Locality-reliability trade-off.} CRANE's image-locality EI (68.1\%) remains below GRACE's 99.5\%, but GRACE's near-perfect EI coexists with $<$1\% rephrase GS; its locality is a byproduct of codebook narrowness, not principled isolation. No baseline achieves strong performance across all dimensions simultaneously: methods with high locality (GRACE, LiveEdit) fail on reliability, while methods with high reliability (FT, LoRA) degrade locality or trigger Structural Collapse. CRANE is the only configuration that maintains high reliability (96.9\% conflict GS) and multi-hop propagation (96.9\% IU) while achieving non-trivial image-locality (68.1\%), with the Cognitive Routing Reward contributing +18pp over SFT-only at negligible reliability cost ($<$1pp).

\textbf{Retrieval bottleneck.} The gold-retrieval experiment on MMEVOKE (87.0\% CEM vs.\ 69.8\% with fusion retrieval) reveals a $\sim$17pp gap attributable to retrieval errors, indicating that retrieval precision is currently the primary bottleneck for OOD deployment. Domain-adaptive retrieval fine-tuning is a promising direction for closing this gap.

\textbf{Model scale.} We validate CRANE on Qwen2.5-VL-7B-Instruct. The framework's design (freezing ViT, training only the LM) is inherently scalable, though hyperparameter adjustments may be needed for larger models.

\textbf{Generality of failure modes.} While Structural Collapse is unique to reasoning MLLMs and Cognitive Dissonance is amplified by them, Shallow Internalization is method-level. Our diagnostic contributions (CoT-aware metrics, ReasonEdit-Bench) are most informative for models with explicit reasoning traces.

%%%%%%%%%%%%%%%%%%%%%%%%%%%%%%%%%%%%%%%%%%%%%%%%%%%%%%%%%%%%%%%%%%%%%%%%%%%%%%%
\section{Conclusion}
\label{sec:conclusion}
%%%%%%%%%%%%%%%%%%%%%%%%%%%%%%%%%%%%%%%%%%%%%%%%%%%%%%%%%%%%%%%%%%%%%%%%%%%%%%%

We have identified three fundamental failure modes of knowledge editing on reasoning MLLMs (Structural Collapse, Cognitive Dissonance, and Shallow Internalization) that are not captured by traditional teacher-forcing and exact-match metrics. Our experiments show that the TF-to-GS gap reaches up to 98.8pp on reasoning MLLMs, that parameter-modification (FT, LoRA) and parameter-isolation (WISE) methods all reduce Vision-R1's format rate to 0\%, while even codebook-based GRACE achieves only 0.5\%, and that GRACE and IKE achieve near-zero multi-hop intermediate entity usage despite moderate rewrite success. We introduced a CoT-aware evaluation protocol backed by LLM-as-a-Judge and constructed ReasonEdit-Bench with conflict stratification, multi-level internalization probes, and multi-hop portability tests to expose these failures.

Our proposed framework, CRANE, mitigates all three failure modes through retrieval-augmented inference (no weight modification), SFT-based structural initialization, and GRPO with a Cognitive Routing Reward. Experiments on ReasonEdit-Bench and the out-of-distribution MMEVOKE benchmark confirm that CRANE maintains strong conflict-scenario reliability and multi-hop knowledge propagation while achieving non-trivial locality, with retrieval precision as the current deployment bottleneck.

These results suggest that effective knowledge editing for reasoning MLLMs benefits from approaches that go beyond surface-level answer injection: training the model to \textit{reason about} edit facts through its explicit chain-of-thought process, rather than merely recalling or parroting them, is one promising direction validated by our experiments.

%%%%%%%%%%%%%%%%%%%%%%%%%%%%%%%%%%%%%%%%%%%%%%%%%%%%%%%%%%%%%%%%%%%%%%%%%%%%%%%
\bibliographystyle{IEEEtran}
\bibliography{references}

\end{document}